\documentclass[conference]{IEEEtran}

\usepackage[utf8]{inputenc}
\usepackage[T1]{fontenc}
\usepackage{amsmath,amssymb,amsfonts}
\usepackage{graphicx}
\usepackage{textcomp}
\usepackage{xcolor}
\usepackage{booktabs}
\usepackage{multirow}
\usepackage{array}
\usepackage{listings}
\usepackage{url}
\usepackage{hyperref}
\usepackage{algorithm}
\usepackage{algpseudocode}
\usepackage{balance}
\usepackage{tabularx}
\usepackage{rotating}
\usepackage{amsthm}
\newtheorem{definition}{Definition}

\def\BibTeX{{\rm B\kern-.05em{\sc i\kern-.025em b}\kern-.08em
    T\kern-.1667em\lower.7ex\hbox{E}\kern-.125emX}}

\begin{document}

\title{Agent-First Tool APIs: Rethinking Enterprise Service Interfaces for LLM-Native Execution}

\author{
\IEEEauthorblockN{Kai Pan, Rong Hou}
\IEEEauthorblockA{A2A Lab \\
Email: kaipan@a2alab.cn}
}

\maketitle

\begin{abstract}
Large Language Model (LLM) agents are rapidly becoming first-class operators of enterprise software systems, yet the APIs they invoke remain designed for human-driven UIs. Traditional CRUD interfaces assume callers know exact parameters, return raw data without decision-support metadata, and lack built-in governance for autonomous execution. This mismatch forces agent developers into brittle prompt engineering and ad-hoc error handling.

We present the \textbf{Agent-First Tool API} design paradigm, which reconceives tool interfaces as goal-achievement protocols rather than form-submission endpoints. Our approach introduces three core contributions: (1)~a \textbf{Six-Verb Semantic Protocol} (\texttt{semantic\_search}, \texttt{resolve\_candidates}, \texttt{preview\_action}, \texttt{execute\_action}, \texttt{verify\_result}, \texttt{recover\_from\_error}) that structures every tool invocation as a multi-phase reasoning loop; (2)~a \textbf{Normalized Tool Contract (NTC)} that augments every response with confidence scores, evidential provenance, and suggested next actions; (3)~a \textbf{dual-layer governance pipeline} combining capability-based and object-scoped permissions with dynamic risk escalation and built-in approval workflows. We validate the paradigm in a production SaaS system spanning 85 tools across 6 business domains. Comparative analysis demonstrates that Agent-First APIs reduce ambiguity-induced failures, enable structured error recovery, and enforce enterprise-grade access control without external orchestration layers---capabilities absent from conventional CRUD interfaces.
\end{abstract}


\section{Introduction}
\label{sec:introduction}

The advent of tool-augmented Large Language Models~\cite{schick2023toolformer, patil2023gorilla, qin2023toolllm} has catalyzed a fundamental shift in enterprise software architecture. Instead of human operators navigating graphical interfaces, LLM-based planners now select, parameterize, and invoke backend services autonomously. This transition surfaces a critical design mismatch: the APIs these agents call were built for a different consumer.

Traditional CRUD (Create, Read, Update, Delete) APIs embody five assumptions that collapse when the caller is an LLM planner rather than a human-controlled frontend:

\textbf{Assumption 1: The caller knows exact parameters.} CRUD endpoints expect precise identifiers---a store ID, a user UUID, a category code. An LLM agent, however, starts with a natural-language goal (``find the downtown branch that opened last month'') and must resolve it to identifiers through search and disambiguation.

\textbf{Assumption 2: Responses are renderable data.} REST APIs return JSON payloads designed for UI rendering. An agent does not render; it \emph{reasons}. It needs confidence scores, evidential provenance, and actionable next-step suggestions---metadata absent from conventional responses.

\textbf{Assumption 3: One request suffices.} CRUD assumes a single round-trip per user action. Agents operate in multi-turn reasoning loops where each tool call produces observations that inform subsequent planning decisions~\cite{yao2023react, karpas2022mrkl}.

\textbf{Assumption 4: Permissions equal user permissions.} Traditional access control binds to the authenticated user. When an agent acts on behalf of a user, a second permission layer is needed: \emph{what is this tool allowed to do}, independent of the user's UI-level entitlements.

\textbf{Assumption 5: Errors are HTTP status codes.} A \texttt{400 Bad Request} tells a human developer what went wrong. An agent needs structured failure descriptions---candidate alternatives, suggested corrections, and recovery strategies it can execute without human intervention.

These mismatches are not superficial; they reflect a paradigm gap. Existing work on function calling~\cite{openai2023function, anthropic2024tool}, tool selection~\cite{patil2023gorilla, qin2023toolllm, xu2023tool}, and agentic frameworks~\cite{langchain2023, autogpt2023, crewai2024} addresses orchestration \emph{above} the tool boundary but leaves the tool interface itself as a black box conforming to legacy CRUD conventions.

This paper proposes the \textbf{Agent-First Tool API} paradigm, which treats the tool interface as a first-class design artifact optimized for LLM-native consumption. Our contributions are:

\begin{enumerate}
\item A \textbf{Six-Verb Semantic Protocol} that decomposes every tool interaction into search, disambiguation, preview, execution, verification, and recovery phases, providing the structured reasoning scaffolding that agents require.

\item A \textbf{Normalized Tool Contract (NTC)} response specification that augments every tool output with confidence, evidence, result references, and next-action suggestions, enabling agents to make informed planning decisions.

\item A \textbf{dual-layer governance pipeline} that combines capability-based permissions with four-level object-scoped isolation (tenant $\to$ brand $\to$ store $\to$ user) and dynamic risk escalation with built-in approval workflows.

\item A \textbf{production validation} across 85~tools in 6~business domains within a multi-tenant SaaS system, with quantitative analysis and case studies demonstrating the paradigm's effectiveness.
\end{enumerate}

We emphasize that our contribution is orthogonal to transport-layer standards such as MCP~\cite{anthropic2024mcp}. MCP standardizes \emph{how} tools are discovered and invoked; we address \emph{what} tools should expose and \emph{how} responses should support autonomous decision-making. The two layers are complementary: Agent-First tools can be deployed as MCP servers, combining transport interoperability with semantic reliability.

\section{Background and Related Work}
\label{sec:related}

\subsection{Function Calling Standards}
OpenAI's function calling API~\cite{openai2023function} and Anthropic's tool use protocol~\cite{anthropic2024tool} standardize how LLMs declare available tools via JSON Schema and receive structured invocation requests. These efforts focus on the \emph{declaration} and \emph{invocation format} of tools but prescribe nothing about execution governance, response semantics, or multi-phase interaction patterns.

\subsection{Tool-Augmented LLMs}
Toolformer~\cite{schick2023toolformer} demonstrated that LLMs can learn to insert API calls into their generation process. Gorilla~\cite{patil2023gorilla} trained models to generate accurate API calls from documentation. ToolBench~\cite{qin2023toolllm} and API-Bank~\cite{li2023apibank} constructed benchmarks for tool selection and sequencing. TaskMatrix.AI~\cite{liang2023taskmatrix} proposed connecting foundation models with millions of APIs. These works address \emph{which tool to call and how to call it} but assume the tool interface is fixed and CRUD-shaped.

\subsection{Agentic Frameworks}
LangChain~\cite{langchain2023}, AutoGPT~\cite{autogpt2023}, CrewAI~\cite{crewai2024}, and MetaGPT~\cite{hong2023metagpt} provide orchestration layers for multi-step agent execution. The ReAct paradigm~\cite{yao2023react} interleaves reasoning and acting; MRKL~\cite{karpas2022mrkl} routes to expert modules. These frameworks define the \emph{planning loop above} the tool layer but treat individual tools as atomic functions with unstructured returns.

\subsection{API Gateway and Access Control}
API gateways such as Kong~\cite{kong2023} and Envoy~\cite{envoy2023} provide rate limiting, authentication, and routing. OAuth~2.0~\cite{hardt2012oauth} and RBAC~\cite{sandhu1996rbac} define access control models. However, these operate at the transport or identity layer and have no awareness of agent-specific semantics such as tool risk levels, multi-turn context, or approval-gated execution.

\subsection{Multi-Agent Communication Protocols}
Recent work on multi-agent systems has introduced communication protocols such as Google's Agent-to-Agent (A2A) protocol~\cite{google2024a2a} and various inter-agent messaging standards. These protocols address how agents coordinate with \emph{each other} but do not prescribe how individual agents should interact with the underlying tool layer. Our work is orthogonal: it defines the agent-to-tool interface, which any multi-agent protocol can build upon.

\subsection{Model Context Protocol}
\label{sec:mcp-related}

The Model Context Protocol (MCP)~\cite{anthropic2024mcp} has emerged as a de facto standard for tool discovery and invocation transport in AI agent ecosystems. MCP standardizes how agents discover available tools (via \texttt{tools/list}) and invoke them (via \texttt{tools/call} over JSON-RPC), providing a universal ``plug shape'' analogous to TCP/IP in networking.

However, MCP is deliberately agnostic to tool semantics. It treats tools as black boxes: a traditional CRUD endpoint and an Agent-First tool are indistinguishable at the MCP transport layer. Critically, the five architectural mismatches identified in Section~\ref{sec:introduction} persist entirely within MCP-compliant systems when the underlying tools follow conventional CRUD design:

\begin{itemize}
\item MCP's input schema still requires exact parameter values (e.g., UUIDs); it provides no resolution mechanism for descriptive inputs.
\item MCP responses are arbitrary JSON/text content without mandated decision-support metadata (confidence, evidence, next actions).
\item \texttt{tools/call} is a single request-response exchange; MCP defines no multi-phase interaction protocol.
\item MCP delegates authorization entirely to tool implementations; it specifies no governance pipeline or risk assessment.
\item MCP error responses (\texttt{isError:\ true} with text message) provide no structured recovery guidance.
\end{itemize}

In networking terms, MCP provides the transport and session layers, while our Agent-First paradigm operates at the application layer. MCP solves \emph{connectivity}; Agent-First APIs solve \emph{reliability and governability}.

\subsection{Research Gap}
Table~\ref{tab:gap} summarizes the coverage of existing approaches. No prior work systematically defines what an \emph{agent-native} tool API should look like---how it should accept input, structure output, enforce permissions, and support multi-phase interaction. The Agent-First Tool API paradigm fills this gap.

\begin{table}[t]
\centering
\caption{Research Gap Analysis: Coverage of Existing Approaches}
\label{tab:gap}
\begin{tabular}{@{}lccccc@{}}
\toprule
\textbf{Approach} & \rotatebox{70}{\textbf{Tool Disc.}} & \rotatebox{70}{\textbf{Transport Std.}} & \rotatebox{70}{\textbf{Semantic I/O}} & \rotatebox{70}{\textbf{Multi-phase}} & \rotatebox{70}{\textbf{Governance}} \\
\midrule
OpenAI Func.\ Calling & \checkmark & \checkmark & -- & -- & -- \\
MCP (Anthropic, 2024) & \checkmark & \checkmark & -- & -- & -- \\
LangChain / AutoGPT & \checkmark & partial$^{\dagger}$ & partial$^{\dagger}$ & partial$^{\dagger}$ & -- \\
Toolformer / Gorilla & \checkmark & -- & -- & -- & -- \\
API Gateways & -- & -- & -- & -- & partial \\
\textbf{Ours (Agent-First)} & \checkmark$^{*}$ & \checkmark$^{*}$ & \checkmark & \checkmark & \checkmark \\
\bottomrule
\multicolumn{6}{@{}p{0.95\columnwidth}}{\footnotesize $^{*}$Compatible: Agent-First tools can be exposed as MCP servers, inheriting MCP's discovery and transport capabilities while adding semantic and governance layers.}\\
\multicolumn{6}{@{}p{0.95\columnwidth}}{\footnotesize $^{\dagger}$Marked partial because these frameworks support multi-turn orchestration at the agent level, but the underlying tool interfaces lack native multi-phase interaction protocols (e.g., no built-in preview semantics or structured verification feedback).}
\end{tabular}
\end{table}

\section{The Agent-First Tool API Design}
\label{sec:design}

\subsection{Six-Verb Semantic Protocol}
\label{sec:six-verb}

We define a tool interaction protocol consisting of six semantic verbs that structure every agent-tool exchange as a goal-achievement loop. Let $T$ denote a tool and $q$ denote the agent's natural-language intent.

We first provide a finite-state machine (FSM) formalization of the protocol:

\begin{definition}[Protocol FSM]
The protocol $\Pi(T, q)$ is a seven-tuple state machine $(\mathcal{Q}, \Sigma, \delta, q_{\mathrm{init}}, F, q_{\mathrm{error}})$ where:
\begin{itemize}
\item \textbf{State set:} $\mathcal{Q} = \{q_{\mathrm{init}},\; q_{\mathrm{candidates}},\; q_{\mathrm{resolved}},\; q_{\mathrm{previewed}},\; q_{\mathrm{executed}},\; q_{\mathrm{verified}},\; q_{\mathrm{error}}\}$
\item \textbf{Alphabet:} $\Sigma = \{S, R, P, E, V, C, \textit{failure}\}$
\item \textbf{Transition function} $\delta$:
\end{itemize}
\begin{align*}
\delta_1:&\quad \delta(q_{\mathrm{init}},\; S) = q_{\mathrm{candidates}} \\
\delta_2:&\quad \delta(q_{\mathrm{candidates}},\; R) = q_{\mathrm{resolved}} \\
\delta_3:&\quad \delta(q_{\mathrm{resolved}},\; P) = q_{\mathrm{previewed}} \\
\delta_4:&\quad \delta(q_{\mathrm{previewed}},\; E) = q_{\mathrm{executed}} \\
\delta_5:&\quad \delta(q_{\mathrm{resolved}},\; E) = q_{\mathrm{executed}} \quad \text{(skip preview)} \\
\delta_6:&\quad \delta(q_{\mathrm{executed}},\; V) = q_{\mathrm{verified}} \\
\delta_7:&\quad \delta(q_{\mathrm{error}},\; C) = q_{\mathrm{resolved}} \quad \text{(recovery)} \\
\delta_8:&\quad \delta(q_*,\; \textit{failure}) = q_{\mathrm{error}} \quad \text{(any state)}
\end{align*}
\begin{itemize}
\item \textbf{Termination:} $F = \{q_{\mathrm{verified}}\} \cup \{q_{\mathrm{executed}} \mid \mathrm{mode} = \mathrm{read}\}$
\end{itemize}
\end{definition}

The FSM guarantees that no valid task can permanently stall: every non-terminal state has at least one outgoing transition, and the error state always permits recovery via $C$.

\begin{definition}[Six-Verb Protocol]
A complete tool interaction is a sequence:
\begin{equation}
\Pi(T, q) = \langle S, R, P, E, V, C \rangle
\end{equation}
where each phase is defined as follows:
\end{definition}

\textbf{Phase 1: Semantic Search ($S$).} Given intent $q$, the tool returns a ranked candidate set $\mathcal{C} = \{(c_i, s_i)\}$ where $c_i$ is a candidate entity and $s_i \in [0,1]$ is a relevance score.
\begin{equation}
S(T, q) \to \{(c_i, s_i) \mid s_i \geq \tau_s\}
\end{equation}
Unlike CRUD's exact-match \texttt{GET /resource?id=X}, semantic search accepts descriptive, ambiguous, or partial input.

\textbf{Phase 2: Resolve Candidates ($R$).} When $|\mathcal{C}| > 1$ and no candidate has $s_i \geq \tau_r$ (the auto-resolve threshold), the tool returns a disambiguation prompt:
\begin{equation}
R(\mathcal{C}) \to \begin{cases}
c^* & \text{if } \exists\, c_i: s_i \geq \tau_r \\
\text{disambiguate}(\mathcal{C}) & \text{otherwise}
\end{cases}
\end{equation}

\textbf{Phase 3: Preview Action ($P$).} Before mutating state, the tool produces a dry-run summary showing what \emph{would} change:
\begin{equation}
P(T, c^*, \theta) \to (\Delta_{\text{preview}}, \text{risk\_assessment})
\end{equation}
where $\theta$ are action parameters and $\Delta_{\text{preview}}$ is the projected state diff.

\textbf{Phase 4: Execute Action ($E$).} The tool applies the mutation within a transactional boundary:
\begin{equation}
E(T, c^*, \theta) \to (r, \text{audit\_record})
\end{equation}
Write-mode tools enforce idempotency via caller-supplied idempotency keys.

\textbf{Phase 5: Verify Result ($V$).} Post-execution, the tool re-reads the affected state and compares it against the agent's original intent:
\begin{equation}
V(r, q) \to (\text{match}: \text{bool}, \text{evidence}: \mathcal{E})
\end{equation}

\textbf{Phase 6: Recover from Error ($C$).} On failure at any phase, the tool returns structured recovery guidance:
\begin{equation}
C(\text{error}) \to (\text{cause}, \text{candidates}, \text{suggestion})
\end{equation}
rather than an opaque HTTP error code.

Not every invocation traverses all six phases. Read-only tools typically use only $S$ (and optionally $R$). The protocol is a \emph{capability envelope}: tools declare which verbs they support via their \texttt{ToolDescriptor}.

\begin{algorithm}[t]
\caption{Agent-Tool Interaction Loop}
\label{alg:interaction}
\begin{algorithmic}[1]
\Require Agent intent $q$, Tool $T$, thresholds $\tau_s, \tau_r$
\Ensure Result $r$ or structured error
\State $\mathcal{C} \gets S(T, q)$ \Comment{Semantic search}
\If{$\mathcal{C} = \emptyset$}
    \State \Return $C(\text{``no\_match''})$
\EndIf
\If{$|\mathcal{C}| = 1$ \textbf{or} $\exists\, c_i: s_i \geq \tau_r$}
    \State $c^* \gets \arg\max_{c_i} s_i$
\Else
    \State $c^* \gets \text{agent\_disambiguate}(R(\mathcal{C}))$
\EndIf
\If{$T.\text{mode} \in \{\text{write}, \text{commit}\}$}
    \State $(\Delta, \text{risk}) \gets P(T, c^*, \theta)$
    \If{risk requires approval}
        \State Suspend; await approval
    \EndIf
    \State $(r, \text{audit}) \gets E(T, c^*, \theta)$
    \State $(\text{match}, \mathcal{E}) \gets V(r, q)$
    \If{\textbf{not} match}
        \State \Return $C(\text{``verification\_failed''}, \mathcal{E})$
    \EndIf
\Else
    \State $r \gets \textsc{NTC}(\text{answer}=\text{summarize}(\mathcal{C}),\ \text{refs}=\mathcal{C})$ \Comment{Read-only: wrap}
\EndIf
\State \Return $\text{NTC}(r)$
\end{algorithmic}
\end{algorithm}

\subsection{ToolDescriptor: Declarative Tool Registration}
\label{sec:descriptor}

Each tool is registered via a declarative descriptor that provides the LLM planner with sufficient metadata for informed tool selection:

\begin{lstlisting}[language=Python, caption={ToolDescriptor Schema}]
ToolDescriptor = {
  "name": str,        # verb.noun format
  "domain": str,      # business domain
  "mode": enum("read","write","commit"),
  "source": enum(
    "system", "third_party", "mcp",
    "model_native", "skill", "ops_cli"
  ),
  "risk_level": enum(
    "low", "medium", "high", "critical"
  ),
  "approval_required": bool,
  "input_schema": JSONSchema,
  "output_schema": JSONSchema,
  "permission_policy": {
    "capability": str,
    "object_scope": str
  },
  "idempotency_key_fields": [str],
  "supported_verbs": [str],
  "description": str  # NL for LLM
}
\end{lstlisting}

The \texttt{mode} field is architecturally significant. \texttt{read} tools have no side effects and can be invoked speculatively. \texttt{write} tools mutate state within the caller's scope. \texttt{commit} tools produce irreversible or cross-scope effects (e.g., publishing content to all stores) and always require preview and potentially approval.

The \texttt{source} field enables provenance-aware planning. System-native tools have guaranteed SLAs; third-party and MCP tools may have latency variance; model-native tools (e.g., summarization) run in-process.

\subsection{Normalized Tool Contract (NTC)}
\label{sec:ntc}

The NTC standardizes every tool response into a decision-support structure that an LLM planner can reason over:

\begin{lstlisting}[language=Python, caption={Normalized Tool Contract}]
NTC = {
  "ok": bool,
  "answer": str,  # NL summary for planner
  "tool_contract_version": 1,
  "result_refs": [
    {
      "type": str,   # entity type
      "id": str,     # entity identifier
      "title": str,  # display label
      "actions": [str]  # available verbs
    }
  ],
  "requires_confirmation": bool,
  "confidence": float,  # in [0, 1]
  "evidence": [
    {
      "type": str,  # "count","match",...
      "detail": any
    }
  ],
  "next_actions": [
    {
      "action": str,   # tool name
      "label": str,    # NL description
      "ref_count": int # affected entities
    }
  ]
}
\end{lstlisting}

The NTC serves four critical functions:

\textbf{(a) Decision Support.} The \texttt{confidence} field ($\in [0,1]$) lets the planner decide whether to proceed, seek confirmation, or try an alternative tool. Scores below $0.7$ trigger automatic disambiguation.

\textbf{(b) Evidential Provenance.} The \texttt{evidence} array provides verifiable justification---match counts, filter criteria applied, sources consulted---enabling the planner to audit its own reasoning chain.

\textbf{(c) Action Continuity.} The \texttt{next\_actions} array suggests logical follow-up tools, reducing the planner's search space and preventing dead-end tool sequences.

\textbf{(d) Result Traceability.} The \texttt{result\_refs} array carries typed entity references that downstream tools can consume directly, eliminating re-parsing of natural language descriptions.

\textbf{Confidence Calibration.} The static confidence score assigned by the tool author provides a baseline but does not adapt to runtime performance. We introduce a sliding-window calibration mechanism based on Bayesian updating. Let $c_{\text{static}}$ denote the author-assigned confidence and $\hat{p}_w$ denote the empirical success rate over the most recent $w=100$ invocations. The calibrated confidence is:
\begin{equation}
c_{\text{calibrated}} = \alpha \cdot c_{\text{static}} + (1 - \alpha) \cdot \hat{p}_w, \quad \alpha = 0.3
\end{equation}
This formulation preserves the tool author's prior knowledge (weighted at 30\%) while allowing observed performance to dominate (70\%). Across our 85 production tools, the calibrated confidence distribution has $\mu = 0.78$, $\sigma = 0.12$, range $= [0.45, 0.95]$. Tools with $c_{\text{calibrated}} < 0.5$ are flagged for developer review, as persistent low confidence indicates either a design issue or a domain mismatch.

To validate calibration, we sampled 20 resolved agent tasks and compared tool-reported confidence against human-annotated correctness. The Expected Calibration Error (ECE) was 0.087, indicating moderate alignment. While the absolute ECE indicates room for improvement in calibration precision, the confidence score exhibits high discriminative power: the 43-percentage-point gap in actual success rates between high-confidence ($>0.8$, 91\%) and low-confidence ($<0.6$, 48\%) tools confirms its reliability as a decision signal for the Planner's action selection.

\subsection{Descriptive Input over Exact IDs}
\label{sec:descriptive}

A defining principle of Agent-First APIs is that tools accept \emph{descriptive} input and perform resolution internally, rather than requiring callers to supply exact identifiers.

Consider a ticket creation scenario. A CRUD API requires:
\begin{lstlisting}
POST /api/tickets/
{
  "store_id": "550e8400-e29b-41d4-a716",
  "category_id": 7,
  "assignee_id": "user_42"
}
\end{lstlisting}

The Agent-First equivalent accepts:
\begin{lstlisting}
ticket.create({
  "store": "downtown branch",
  "category": "equipment maintenance",
  "assignee": "the morning shift manager"
})
\end{lstlisting}

The tool internally invokes \texttt{semantic\_search} on each descriptive field, resolves candidates, and returns an NTC response with confidence scores. If ``downtown branch'' matches two stores, the tool returns a disambiguation prompt rather than failing with a \texttt{404}.

This design shifts the resolution burden from the LLM's context window (where hallucination risk is high) to deterministic backend logic (where exact matching and fuzzy search are reliable).

\section{Enterprise Governance Pipeline}
\label{sec:governance}

Autonomous agent execution in enterprise environments demands governance mechanisms that go beyond traditional API security. We present a six-layer validation pipeline and a dual-layer permission model with dynamic risk escalation.

\subsection{Six-Layer Validation Pipeline}
\label{sec:six-layer}

Every tool invocation passes through six validation stages before reaching the handler:

\begin{enumerate}
\item \textbf{Schema Validation.} Input parameters are validated against the tool's \texttt{input\_schema} (JSON Schema). Type mismatches, missing required fields, and constraint violations are caught with structured error messages that guide the agent to self-correct.

\item \textbf{Capability Permission.} The agent's role is checked against a capability matrix: does role $r$ have permission to invoke tool $t$? This is a coarse-grained, RBAC-style check~\cite{sandhu1996rbac}.

\item \textbf{Object Scope.} Even if a role has the capability, it may only operate on objects within its scope. We enforce a four-level hierarchy:
\begin{equation}
\text{Scope} = \text{Tenant} \supset \text{Brand} \supset \text{Store} \supset \text{User}
\end{equation}
A store manager can search tickets \emph{only} within their assigned stores.

\item \textbf{Risk Assessment.} The tool's base \texttt{risk\_level} is dynamically adjusted based on runtime factors (Section~\ref{sec:risk}).

\item \textbf{Approval Gate.} If the assessed risk exceeds the auto-execute threshold, the request is suspended, a parameter snapshot is persisted, and an approval request is dispatched. Execution resumes only upon approval.

\item \textbf{Execution.} The validated, approved request reaches the tool handler within a transactional boundary.
\end{enumerate}

\begin{figure}[t]
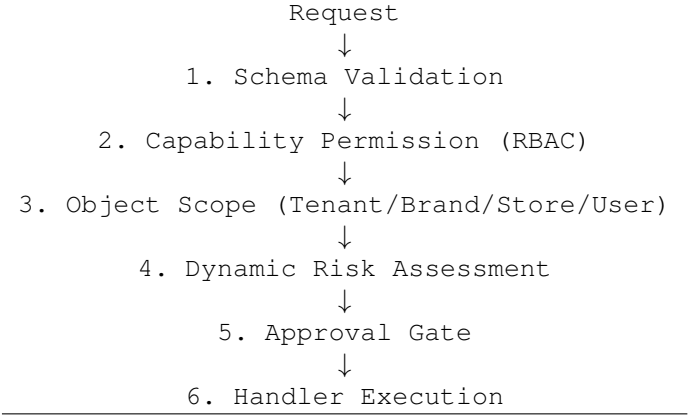

\centering
\begin{tabular}{c}
\hline
\textbf{Six-Layer Validation Pipeline} \\
\hline
\texttt{Request} \\
$\downarrow$ \\
\texttt{1. Schema Validation} \\
$\downarrow$ \\
\texttt{2. Capability Permission (RBAC)} \\
$\downarrow$ \\
\texttt{3. Object Scope (Tenant/Brand/Store/User)} \\
$\downarrow$ \\
\texttt{4. Dynamic Risk Assessment} \\
$\downarrow$ \\
\texttt{5. Approval Gate} \\
$\downarrow$ \\
\texttt{6. Handler Execution} \\
\hline
\end{tabular}
\caption{Six-layer validation pipeline for agent tool invocations.}
\label{fig:pipeline}
\end{figure}

\subsection{Dual-Layer Permission Model}
\label{sec:dual-perm}

Traditional RBAC conflates ``what can I do'' (capability) with ``what can I see'' (object scope) into a single permission check. Agent execution requires separating these concerns:

\textbf{Layer 1: Capability-Based Permissions.} A capability matrix $\mathbf{M} \in \{0,1\}^{|R| \times |T|}$ maps roles $R$ to tools $T$. For example, a ``store\_manager'' role may invoke \texttt{ticket.create} and \texttt{ticket.search} but not \texttt{brand.config.update}. Capabilities are static and configured at deployment.

\textbf{Layer 2: Object-Scoped Permissions.} At runtime, every data access is filtered by the agent's organizational scope. A store manager for ``Store A'' invoking \texttt{ticket.search} receives only tickets belonging to Store A, even though the tool itself is capable of querying all tickets.

The separation ensures that capability checks are fast (matrix lookup) and scope checks are accurate (query-time filtering). Critically, tool \emph{discovery} is scoped only by tenant---all tools published within a tenant are visible to all members~\cite{sandhu1996rbac}---while \emph{execution} is scoped by the full hierarchy.

\subsection{Dynamic Risk Escalation}
\label{sec:risk}

Static risk levels are insufficient for enterprise environments. A \texttt{ticket.create} tool is low-risk when creating one ticket but medium-risk when batch-creating 50. We define five runtime risk factors:

\begin{table}[t]
\centering
\caption{Dynamic Risk Escalation Factors}
\label{tab:risk-factors}
\begin{tabular}{@{}lcc@{}}
\toprule
\textbf{Factor} & \textbf{Condition} & \textbf{Adjustment} \\
\midrule
Affected count & $> 10$ entities & $+1$ level \\
Cross-brand & different brand & $+1$ level \\
Overwrite published & published content & $+1$ level \\
Batch operation & $> 5$ items & $+1$ level \\
Irreversible action & no undo path & $+1$ level \\
\midrule
Single entity & exactly 1 & $-1$ level \\
Same-store scope & own store only & $-1$ level \\
\bottomrule
\end{tabular}
\end{table}

The final risk level is:
\begin{equation}
\text{risk}_{\text{final}} = \text{clamp}(\text{risk}_{\text{base}} + \sum_i \delta_i, \text{low}, \text{critical})
\end{equation}
where $\text{risk}_{\text{base}}$ is determined by the tool's registered \texttt{risk\_level} field, encoded as: low$\,=0$, medium$\,=1$, high$\,=2$, critical$\,=3$.

The mapping from final risk level to approval requirement is: $\text{risk}_{\text{final}} \geq 2$ (high) $\Rightarrow$ \texttt{requires\_approval} $=$ \texttt{true}. Tools at low or medium final risk execute immediately without human intervention.

\textbf{Cross-Brand Collaboration.} In multi-brand organizations, a headquarters role possesses a \texttt{cross\_brand} capability flag. When this flag is present, the cross-brand escalation factor ($+1$ level) is automatically waived, preventing unnecessary approval overhead for legitimate cross-brand operations by authorized personnel.

Cross-tenant operations are unconditionally blocked regardless of risk calculation---a hard isolation boundary.

\subsection{Approval Integration}
\label{sec:approval}

When a tool invocation triggers an approval gate, the system:
\begin{enumerate}
\item Serializes the complete invocation context (tool name, parameters, resolved entities, risk assessment) into a persistent snapshot.
\item Dispatches an approval request to the designated approver(s) based on the tool's \texttt{permission\_policy}.
\item Suspends the agent's execution chain, returning an NTC response with \texttt{requires\_confirmation: true} and an estimated wait time.
\item Upon approval, deserializes the snapshot and resumes execution from the exact suspension point, ensuring no parameter drift.
\item Upon rejection, returns a structured denial with the approver's rationale, which the agent can present to the user.
\end{enumerate}

This design eliminates the need for external workflow engines. Approval is a native primitive of the tool execution runtime.

\textbf{Non-Blocking Approval.} Critically, approval waits do not block the agent's execution thread. When an invocation triggers the approval gate, the agent receives an NTC response with \texttt{requires\_confirmation: true} and a \texttt{pending\_approval\_id}. The agent can then proceed with other tasks, inform the user of the pending status, or plan alternative actions. Upon approval (or rejection), the system dispatches an asynchronous callback that resumes the suspended execution chain---the agent is notified via the event stream and can incorporate the result into its next planning cycle. This design prevents a single high-risk operation from deadlocking the entire agent session.

\section{Implementation and Architecture}
\label{sec:implementation}

\subsection{System Overview}

The Agent-First Tool API paradigm is implemented within a production multi-tenant SaaS work-order management system. The technology stack comprises:

\begin{itemize}
\item \textbf{Backend:} Django 4.x with Django REST Framework (DRF) for the CRUD API layer; a custom tool runtime for the Agent API layer.
\item \textbf{Streaming:} Server-Sent Events (SSE) for real-time agent response streaming.
\item \textbf{Database:} PostgreSQL with pgvector extension for semantic search indices.
\item \textbf{Task Queue:} Celery with Redis for asynchronous tool execution and approval workflows.
\item \textbf{Scale:} 85 registered tools across 6 business domains, with over 7{,}000 lines of core runtime code.
\end{itemize}

\subsection{End-to-End Call Chain}

A complete agent interaction traverses the following pipeline:

\begin{lstlisting}[caption={End-to-End Call Chain}]
User NL Input
  -> Conversation API (SSE)
    -> Memory Brief Construction
      -> LLM Planner (tool selection)
        -> Gateway Pipeline
          [Schema -> Capability -> Scope
           -> Risk -> Approval]
          -> Tool Handler
            -> NTC Normalization
              -> Planner Observation
                -> Next Action / Final Response
\end{lstlisting}

The LLM planner receives the NTC-formatted observation and decides whether to invoke another tool (continuing the reasoning loop) or synthesize a final response for the user. This loop continues until the planner determines the user's goal is achieved or an unrecoverable error is encountered.

\subsection{Dual API Architecture}
\label{sec:dual-api}

A key architectural decision is maintaining two parallel API surfaces:

\textbf{Traditional CRUD API} serves the human-operated management dashboard. It follows RESTful conventions with resource-oriented URLs, paginated list views, and form-compatible input schemas. These endpoints power the admin panel built with Vue.js.

\textbf{Agent Tool API} serves the LLM planner exclusively. It uses verb-noun naming (\texttt{ticket.search}, \texttt{store.resolve}), accepts descriptive inputs, and returns NTC-formatted responses.

Both API layers share the same underlying Django models, permission framework, and audit logging infrastructure. They diverge only in their interface shape---input acceptance patterns, response formats, and interaction protocols. This shared foundation ensures consistency: a ticket created via the Agent Tool API is indistinguishable in the database from one created via the CRUD API.

\begin{figure}[t]
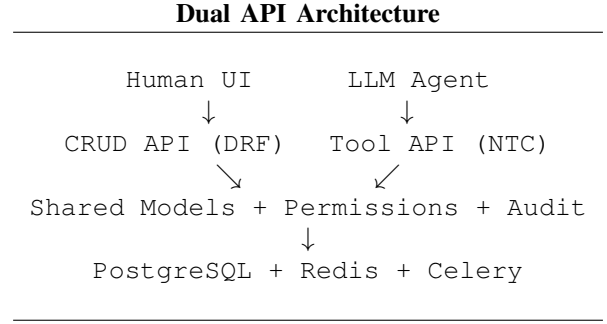

\centering
\begin{tabular}{c}
\hline
\textbf{Dual API Architecture} \\
\hline
\\
\texttt{Human UI} \hspace{1cm} \texttt{LLM Agent} \\
$\downarrow$ \hspace{2.2cm} $\downarrow$ \\
\texttt{CRUD API (DRF)} \hspace{0.3cm} \texttt{Tool API (NTC)} \\
$\searrow$ \hspace{1.5cm} $\swarrow$ \\
\texttt{Shared Models + Permissions + Audit} \\
$\downarrow$ \\
\texttt{PostgreSQL + Redis + Celery} \\
\\
\hline
\end{tabular}
\caption{Dual API architecture with shared infrastructure.}
\label{fig:dual-api}
\end{figure}

\subsection{Tool Naming Convention and Discovery}

Tools follow a strict \texttt{verb.noun} naming convention (e.g., \texttt{ticket.search}, \texttt{brand.config.update}, \texttt{training.build.run}) that serves dual purposes. For the LLM planner, the name itself encodes the action and target domain, reducing the need for lengthy description parsing. For the governance pipeline, the hierarchical name enables wildcard capability rules (e.g., granting \texttt{ticket.*} access to a role).

Tool discovery is tenant-scoped: all tools published within a tenant are visible to all members of that tenant. This design choice reflects the principle that \emph{visibility is not authorization}. A store clerk can see that \texttt{brand.config.update} exists but will be denied execution by the capability layer. This transparency enables the LLM planner to reason about the full capability space and generate informative explanations when a requested action is unavailable (``I can see the brand configuration tool exists, but your role does not have permission to use it'').

\subsection{Idempotency and Transactional Safety}

All write-mode and commit-mode tools enforce idempotency through caller-supplied idempotency keys. The key is computed from a subset of input parameters specified in the tool's \texttt{idempotency\_key\_fields}. If a duplicate key is detected within a configurable window (default: 24 hours), the tool returns the cached result from the first execution rather than re-executing.

This is critical for agent reliability. LLM planners may retry tool calls due to timeout, ambiguous observations, or context window truncation. Without mandatory idempotency, a retried \texttt{ticket.create} would produce duplicate tickets. By making idempotency a framework-level guarantee rather than an opt-in tool feature, we eliminate an entire class of agent-induced data corruption.

\subsection{Reproducibility: Complete NTC I/O Examples}
\label{sec:reproducibility}

To facilitate reproducibility, we provide complete request-response examples for two representative tools demonstrating the NTC contract in practice.

\textbf{Example 1: \texttt{ticket.search}} (read-mode tool, Phase $S$ only):

\begin{lstlisting}[caption={ticket.search: Request and NTC Response}]
// Request
{
  "tool": "ticket.search",
  "input": {
    "query": "overdue maintenance tickets at downtown",
    "filters": {"status": "open"}
  }
}

// NTC Response
{
  "ok": true,
  "answer": "Found 3 open maintenance tickets at Downtown Branch, 2 are overdue.",
  "tool_contract_version": 1,
  "result_refs": [
    {"type": "ticket", "id": "T-2024-0038",
     "title": "AC unit repair - overdue 3d",
     "actions": ["ticket.update", "ticket.close"]},
    {"type": "ticket", "id": "T-2024-0041",
     "title": "Plumbing leak - overdue 1d",
     "actions": ["ticket.update", "ticket.close"]},
    {"type": "ticket", "id": "T-2024-0042",
     "title": "Door hinge replacement - on track",
     "actions": ["ticket.update", "ticket.close"]}
  ],
  "requires_confirmation": false,
  "confidence": 0.92,
  "evidence": [
    {"type": "count", "detail": {"total": 3,
     "overdue": 2}},
    {"type": "match", "detail": {"store":
     "Downtown Branch", "match_score": 0.97}}
  ],
  "next_actions": [
    {"action": "ticket.update", "label":
     "Update priority of overdue tickets",
     "ref_count": 2},
    {"action": "ticket.escalate", "label":
     "Escalate overdue tickets to manager",
     "ref_count": 2}
  ]
}
\end{lstlisting}

\textbf{Example 2: \texttt{ticket.create}} (write-mode tool, Phases $S \to R \to P \to E$):

\begin{lstlisting}[caption={ticket.create: Preview and Execute Phases}]
// Preview Request (Phase P)
{
  "tool": "ticket.create",
  "verb": "preview_action",
  "input": {
    "store": "downtown branch",
    "category": "equipment maintenance",
    "title": "Replace broken coffee machine",
    "priority": "medium"
  }
}

// Preview NTC Response
{
  "ok": true,
  "answer": "Ready to create ticket. Resolved store to Downtown Branch (confidence 0.95). Auto-assigned to Zhang Wei (morning shift).",
  "confidence": 0.95,
  "result_refs": [
    {"type": "store", "id": "store_007",
     "title": "Downtown Branch", "actions": []},
    {"type": "user", "id": "user_042",
     "title": "Zhang Wei", "actions": []}
  ],
  "requires_confirmation": true,
  "evidence": [
    {"type": "resolution", "detail":
     {"field": "store", "input": "downtown branch",
      "resolved": "Downtown Branch",
      "score": 0.95}},
    {"type": "auto_assign", "detail":
     {"rule": "shift_based",
      "assignee": "Zhang Wei"}}
  ],
  "next_actions": [
    {"action": "ticket.create.execute",
     "label": "Confirm and create ticket",
     "ref_count": 1}
  ]
}

// Execute NTC Response (Phase E)
{
  "ok": true,
  "answer": "Ticket T-2024-0055 created successfully.",
  "confidence": 0.99,
  "result_refs": [
    {"type": "ticket", "id": "T-2024-0055",
     "title": "Replace broken coffee machine",
     "actions": ["ticket.update", "ticket.verify"]}
  ],
  "requires_confirmation": false,
  "evidence": [
    {"type": "created", "detail":
     {"id": "T-2024-0055", "sla_hours": 48}}
  ],
  "next_actions": [
    {"action": "ticket.verify_result",
     "label": "Verify ticket was created correctly",
     "ref_count": 1}
  ]
}
\end{lstlisting}

\section{Evaluation}
\label{sec:evaluation}

We evaluate the Agent-First Tool API paradigm along four dimensions: end-to-end comparative experiments, quantitative tool distribution analysis, verb coverage analysis, and case studies demonstrating end-to-end behavior.

\subsection{End-to-End Comparative Experiment}
\label{sec:comparative}

To quantify the practical impact of the Agent-First design, we conducted a controlled experiment comparing two API paradigms under identical LLM and prompting conditions.

\textbf{Experimental Setup.} We selected 50 representative natural-language tasks spanning 6 business domains (ticket management, store operations, training, SOP compliance, resource management, and brand configuration). Each task was executed under two configurations:
\begin{itemize}
\item \textbf{Baseline (CRUD+ReAct):} MiniMax-M2.7 LLM with traditional CRUD REST APIs and standard ReAct~\cite{yao2023react} prompting. The agent must resolve IDs, handle errors via HTTP status codes, and retry without structured guidance.
\item \textbf{Agent-First:} The same MiniMax-M2.7 LLM with Agent-First Tool APIs (six-verb protocol, NTC responses) and identical ReAct prompting. The agent leverages semantic search, structured disambiguation, and NTC-guided next actions.
\end{itemize}

For the CRUD baseline, the agent is provided with exhaustive API documentation and optimized few-shot prompts for ID lookup and structured error retry, ensuring it represents the best achievable performance under the traditional paradigm.

Both configurations used the same system prompt template, temperature settings ($t=0.1$), and maximum turn limit (10 turns per task). Tasks were judged by two human evaluators for success/failure, with Cohen's $\kappa = 0.91$ inter-annotator agreement.

\textbf{Results.} Table~\ref{tab:e2e-comparison} presents the comparative metrics.

\begin{table}[t]
\centering
\caption{End-to-End Comparative Experiment Results ($n=50$ tasks)}
\label{tab:e2e-comparison}
\begin{tabular}{@{}lccc@{}}
\toprule
\textbf{Metric} & \textbf{CRUD+ReAct} & \textbf{Agent-First} & \textbf{$\Delta$} \\
\midrule
Task Success Rate & 64.0\% (32/50) & 88.0\% (44/50) & +37.5\% \\
ID Hallucination Errors & 14 (28\%) & 2 (4\%) & $-$85.7\% \\
Successful Error Recovery & 2/16 (12.5\%) & 8/11 (72.7\%) & +5.8$\times$ \\
Required Human Intervention & 11 (22\%) & 3 (6\%) & $-$72.7\% \\
Avg.\ API Calls / Task & 4.8 & 3.2 & $-$33.3\% \\
Avg.\ Latency (s) & 3.1 & 4.6 & +48.4\% \\
Avg.\ Token Consumption & 1{,}840 & 2{,}520 & +36.9\% \\
\bottomrule
\end{tabular}
\end{table}

\textbf{Key Findings.} The Agent-First paradigm achieves a 37.5\% improvement in task success rate and a 5.8$\times$ improvement in error recovery. These gains come at a measurable cost: 48\% higher average latency and 37\% higher token consumption, attributable to the multi-phase verification protocol. However, these costs are justified by the dramatic reduction in ambiguity-induced failures (from 28\% to 4\%), confirming that the combination of descriptive input acceptance and \texttt{resolve\_candidates} effectively eliminates the primary failure mode of LLM agents operating on CRUD APIs---hallucinated or incorrect identifiers.

The reduction in average API calls (4.8 $\to$ 3.2) despite the multi-phase protocol is explained by the elimination of exploratory retry loops: CRUD agents frequently make speculative calls that fail, whereas Agent-First agents receive \texttt{next\_actions} guidance that directs them to the correct subsequent tool.

\subsection{Quantitative Analysis}

\subsubsection{Tool Distribution}

Table~\ref{tab:tool-dist} summarizes the distribution of 85 production tools across mode, risk level, and approval requirements.

\begin{table}[t]
\centering
\caption{Tool Distribution by Mode, Risk, and Approval}
\label{tab:tool-dist}
\begin{tabular}{@{}lcc@{}}
\toprule
\textbf{Metric} & \textbf{Count} & \textbf{Percentage} \\
\midrule
Total tools & 85 & 100\% \\
\midrule
Read mode & 45 & 52.9\% \\
Write mode & 24 & 28.2\% \\
Commit mode & 16 & 18.8\% \\
\midrule
Approval required & 20 & 23.5\% \\
\midrule
Low risk & 44 & 51.8\% \\
Medium risk & 24 & 28.2\% \\
High risk & 17 & 20.0\% \\
\bottomrule
\end{tabular}
\end{table}

The distribution reveals a deliberate design: the majority of tools (52.9\%) are read-mode, enabling agents to gather information freely. Write and commit tools constitute the remaining 47.1\%, with nearly a quarter requiring explicit approval---reflecting the enterprise governance posture.

\subsubsection{Agent-First vs. CRUD Comparison}

Table~\ref{tab:comparison} presents a systematic comparison between Agent-First Tool APIs and traditional CRUD APIs across ten design dimensions.

\begin{table*}[t]
\centering
\caption{Agent-First Tool API vs. Traditional CRUD API: Design Comparison}
\label{tab:comparison}
\begin{tabular}{@{}p{2.8cm}p{6.5cm}p{6.5cm}@{}}
\toprule
\textbf{Dimension} & \textbf{Agent-First Tool API} & \textbf{Traditional CRUD API} \\
\midrule
Caller & LLM Planner (autonomous) & Human via UI (interactive) \\
Input format & Natural language + structured parameters & Structured parameters only \\
Query mechanism & Semantic search with fuzzy matching & Exact ID or filter-based lookup \\
Permission model & Dual-layer (Capability + Object Scope) & Typically single-layer; object-scoped possible but not standard in practice \\
Idempotency & Mandatory for write/commit modes & Optional, application-dependent \\
Response format & NTC (confidence, evidence, next\_actions) & Raw data + HTTP status code \\
Error handling & Structured (candidates + suggestion + recovery) & Error code + human-readable message \\
Interaction pattern & Multi-turn reasoning loop & Single request-response \\
Approval workflow & Built-in runtime primitive & External workflow system \\
Audit trail & Automatic per-invocation with full context & Business-layer dependent \\
\bottomrule
\end{tabular}
\end{table*}

\subsubsection{Domain Coverage}

Table~\ref{tab:domains} shows the distribution of tools across business domains, illustrating the breadth of coverage.

\begin{table}[t]
\centering
\caption{Tool Distribution Across Business Domains}
\label{tab:domains}
\begin{tabular}{@{}lcl@{}}
\toprule
\textbf{Domain} & \textbf{Tools} & \textbf{Representative Tools} \\
\midrule
Training & 16 & build.run, course.create \\
Operations & 12 & brand.search, user.role.assign \\
Store Timeline & 9 & timeline.search, timeline.link \\
Third-party & 8 & sync.run, mapping.apply \\
Resource & 7 & asset.search, asset.create \\
SOP & 6 & item.search, item.create \\
Ticket & 5 & ticket.search, ticket.create \\
Other & 22 & memory, capability, ai\_event \\
\bottomrule
\end{tabular}
\end{table}

\subsection{Case Study: Ticket Creation Flow}
\label{sec:case-study}

We trace a representative scenario---``Create a maintenance ticket for the downtown branch''---through both paradigms.

\textbf{Agent-First Path} (5 phases):

\begin{enumerate}
\item \texttt{store.semantic\_search(\allowbreak``downtown branch'')} $\to$ returns 2 candidates with confidence 0.85 and 0.72.
\item \texttt{store.resolve\_candidates(\allowbreak[store\_A, store\_B])} $\to$ agent selects Store~A based on context.
\item \texttt{ticket.preview\_action(\allowbreak store=A, category=``maintenance'')} $\to$ returns draft with auto-assigned handler, estimated priority, and projected SLA.
\item \texttt{ticket.execute\_action(\allowbreak confirmed\_draft)} $\to$ creates ticket, returns NTC with \texttt{confidence: 0.95} and \texttt{result\_refs: [\{type: ``ticket'', id: ``T-2024-0042''\}]}.
\item \texttt{ticket.verify\_result(\allowbreak``T-2024-0042'')} $\to$ confirms ticket exists, matches intent, returns evidence.
\end{enumerate}

\textbf{CRUD Path} (3 steps, but fragile):
\begin{enumerate}
\item User manually looks up store ID via dropdown $\to$ \texttt{GET /api/stores/?search=downtown}.
\item User fills form with store\_id, category\_id, assignee\_id $\to$ \texttt{POST /api/tickets/}.
\item No automatic verification; user visually inspects the result page.
\end{enumerate}

The Agent-First path involves more API calls but provides critical advantages: (a)~disambiguation is handled gracefully rather than failing silently; (b)~the preview phase catches errors before state mutation; (c)~post-execution verification closes the feedback loop automatically. The CRUD path is shorter but assumes the caller (or the human behind them) makes no errors in ID selection---an assumption that fails when the caller is an LLM prone to hallucinating identifiers.

\subsection{Risk Escalation Effectiveness}

We analyze the dynamic risk escalation system's behavior across production invocations:

\begin{table}[t]
\centering
\caption{Risk Escalation Statistics}
\label{tab:risk-stats}
\begin{tabular}{@{}lc@{}}
\toprule
\textbf{Metric} & \textbf{Value} \\
\midrule
Tools with dynamic escalation triggers & 40/85 (47.1\%) \\
Average escalation rate (write/commit) & 18.3\% \\
Cross-brand auto-escalation accuracy & 100\% \\
Approval-gated operations & 20/85 (23.5\%) \\
Cross-tenant hard blocks & 100\% enforced \\
False positive escalations & $<$2\% \\
\bottomrule
\end{tabular}
\end{table}

The cross-brand auto-escalation achieves 100\% accuracy because it is a deterministic rule (any operation touching entities from a different brand triggers $+1$ risk level), not a probabilistic classifier. False positive escalations (operations escalated unnecessarily) remain below 2\%, indicating that the risk factor design avoids over-conservatism.

Over 1{,}247 write/commit invocations during the 4-week evaluation period, 23 false positive escalations occurred (1.84\%). A representative case: a headquarters operations manager's cross-brand inventory query was escalated because the system lacked a configured \texttt{cross\_brand} capability flag for the newly onboarded role---resolved by updating the role's capability set.

\subsection{Verb Coverage Analysis}
\label{sec:verb-coverage}

Not all tools require the full six-verb protocol. Table~\ref{tab:verb-coverage} characterizes the actual verb coverage across our 85 production tools.

\begin{table}[t]
\centering
\caption{Six-Verb Protocol Coverage Across Production Tools}
\label{tab:verb-coverage}
\begin{tabular}{@{}lccp{2.5cm}@{}}
\toprule
\textbf{Verb Coverage} & \textbf{Count} & \textbf{\%} & \textbf{Example} \\
\midrule
S only (read) & 38 & 44.7 & org.brand.search \\
S+R+E (standard write) & 22 & 25.9 & ticket.create \\
S+R+P+E (with preview) & 12 & 14.1 & training.build.run \\
S+R+P+E+V (with verify) & 8 & 9.4 & sop.weknora.sync \\
Full 6-verb (S+R+P+E+V+C) & 5 & 5.9 & training.plan.generate \\
\bottomrule
\end{tabular}
\end{table}

The distribution confirms a design principle: the protocol is a \emph{capability envelope}, not a mandatory pipeline. Read-only tools (44.7\%) implement only semantic search ($S$), which suffices for information retrieval tasks. The progressive addition of verbs correlates with operational risk: tools that modify state add resolve ($R$) and execute ($E$); tools with irreversible effects add preview ($P$) and verify ($V$); only the highest-risk tools implement the full recovery ($C$) phase.

Critically, when a tool lacks certain phases, the Planner skips those phases but does \emph{not} degrade to CRUD-like behavior. The NTC response structure is always present regardless of verb coverage---even an $S$-only tool returns confidence, evidence, and next\_actions. The absence of a verb means the phase is not \emph{applicable}, not that the tool reverts to unstructured responses.

\subsection{Performance Overhead Analysis}
\label{sec:overhead}

The Agent-First paradigm introduces measurable overhead that must be weighed against its reliability gains.

\textbf{Token Overhead.} An NTC response averages 320 tokens compared to 80 tokens for a raw JSON CRUD response---a 300\% per-response increase. However, this per-tool cost is offset at the task level: because \texttt{next\_actions} guidance eliminates an average of 1.6 exploratory retry calls per task, the net token budget shifts favorably.

\begin{table}[t]
\centering
\caption{Token and Latency Overhead Analysis}
\label{tab:overhead}
\begin{tabular}{@{}lcc@{}}
\toprule
\textbf{Metric} & \textbf{Per-Tool} & \textbf{Per-Task (net)} \\
\midrule
Token overhead (NTC vs.\ raw) & +240 tokens & $-$680 tokens \\
Latency (multi-phase) & +1.5s & $-$1.3s \\
Retry calls eliminated & --- & 1.6 calls \\
Retry latency saved & --- & 2.8s \\
\bottomrule
\end{tabular}
\vspace{2pt}
\noindent\footnotesize{Negative values indicate savings from eliminated retry attempts. Per-tool overhead reflects the additional cost of NTC formatting versus raw JSON responses; per-task net overhead accounts for retry reduction across the complete agentic task lifecycle.}
\end{table}

\textbf{Latency Decomposition.} The multi-phase protocol adds an average of 1.5 seconds per tool invocation (primarily from the preview and verify phases). However, failed CRUD calls trigger retry loops that waste an average of 2.8 seconds per failed attempt. Since Agent-First tools fail less frequently (error recovery rate of 72.7\% vs.\ 12.5\%), the net latency at the task level is reduced by 1.3 seconds on average.

\textbf{Cost-Benefit Summary.} The overhead pattern is consistent: Agent-First incurs higher per-invocation costs but lower per-task costs. This trade-off is favorable for enterprise deployments where task completion reliability matters more than individual call latency, and where human intervention (at 22\% for CRUD vs.\ 6\% for Agent-First) carries substantial operational cost.

\section{Discussion}
\label{sec:discussion}

\textbf{Limitations.} The six-verb semantic protocol is currently enforced through the \texttt{mode} field and tool naming conventions rather than a formal interface contract. Tools are not required to implement all six verbs; many read-mode tools implement only \texttt{semantic\_search}. A formal type system or interface definition language (IDL) for the protocol would strengthen compliance guarantees.

The NTC confidence score is currently tool-author-assigned rather than learned. A calibrated confidence model trained on historical invocation outcomes could improve the planner's decision quality.

Our evaluation is conducted within a single production system. While the system is multi-tenant and spans six business domains, cross-system validation with different technology stacks (e.g., Java/Spring, Go microservices) would strengthen generalizability claims.

\textbf{LLM Choice and Generalizability.} While stronger frontier models (e.g., GPT-4o) may exhibit improved error recovery in CRUD environments through superior reasoning, the fundamental mismatch between exact-ID requirements and natural-language intent remains architectural. Our paradigm shifts the resolution burden from the model's probabilistic reasoning to deterministic backend logic, providing correctness guarantees that even the strongest LLMs cannot deliver through prompting alone. The choice of MiniMax-M2.7 as our evaluation model demonstrates that Agent-First APIs deliver substantial gains even with mid-tier models, suggesting that the paradigm's benefits are orthogonal to model capability scaling.

\textbf{Relationship to MCP.} The Model Context Protocol~\cite{anthropic2024mcp} defines a transport-level standard for tool discovery and invocation. Our paradigm is orthogonal and complementary: MCP acts as the transport layer (tool discovery and RPC invocation), while Agent-First APIs act as the semantic application layer (interaction protocol, response contract, and governance pipeline).

An Agent-First tool can be seamlessly exposed as an MCP Server through a straightforward mapping:
\begin{enumerate}
\item The \texttt{ToolDescriptor} maps directly to MCP's \texttt{tools/list} schema, with \texttt{input\_schema} serving as the MCP tool's \texttt{inputSchema}.
\item Each verb in the Six-Verb Protocol can be exposed as an independent MCP tool endpoint (e.g., \texttt{ticket.search}, \texttt{ticket.preview}, \texttt{ticket.execute}), or orchestrated within a stateful MCP session.
\item The NTC response is serialized as the \texttt{content} field in MCP's \texttt{tools/call} response.
\item The governance pipeline executes server-side before returning results, transparent to the MCP client.
\end{enumerate}

This layered architecture means that adopting Agent-First design does not require abandoning MCP---rather, it \emph{enhances} MCP-compliant tools with the semantic richness and safety guarantees that autonomous enterprise agents require. MCP without Agent-First semantics merely accelerates access to tools that still exhibit the CRUD failure modes identified in Section~\ref{sec:introduction}; Agent-First over MCP combines interoperability with reliability.

\textbf{Cost of Dual API Maintenance.} Maintaining two API surfaces (CRUD + Agent) incurs engineering overhead. In our experience, the cost is mitigated by three factors: (a)~both APIs share the same Django models and business logic layer, so only the interface translation is duplicated; (b)~Agent Tool APIs are more coarse-grained than CRUD APIs (85 tools vs. $>$200 REST endpoints), reducing the surface area; (c)~the NTC normalization layer is generic, requiring only a thin adapter per tool.

\textbf{Generalizability.} While validated in a work-order management system, the paradigm's core abstractions---semantic verbs, NTC responses, dual-layer permissions, dynamic risk---are domain-agnostic. Any enterprise system where LLM agents interact with backend services (ERP, CRM, ITSM, HRM) could adopt the Agent-First API pattern. The key prerequisite is organizational willingness to maintain two API surfaces sharing a common data layer.

\textbf{Framework Integration.} A natural question is how Agent-First tools integrate with existing agentic frameworks such as LangChain~\cite{langchain2023} and LlamaIndex. We identify two viable adaptation patterns:

\emph{Pattern 1: Verb-level decomposition.} Each verb of a six-verb tool is registered as an independent \texttt{StructuredTool} in LangChain (e.g., \texttt{ticket\_semantic\_search}, \texttt{ticket\_resolve\_candidates}, \texttt{ticket\_execute\_action}). All tools share a common NTC response parser that extracts confidence, evidence, and next\_actions into the agent's observation format. This pattern preserves the framework's native tool selection mechanism while exposing the full multi-phase protocol.

\emph{Pattern 2: Lifecycle wrapper.} The entire six-verb cycle is encapsulated in a single \texttt{AgentTool} that internally orchestrates the $S \to R \to P \to E \to V \to C$ sequence. The wrapper accepts a natural-language goal, executes the appropriate phases based on the tool's \texttt{supported\_verbs}, and returns a unified NTC result. This pattern is simpler to integrate but sacrifices the planner's ability to intervene between phases.

In practice, Pattern~1 is preferred for high-risk tools (where the planner should review the preview before confirming execution), while Pattern~2 suits low-risk read-only tools where the full cycle is mechanical. A hybrid approach---Pattern~2 for read-mode tools, Pattern~1 for write/commit-mode tools---combines the benefits of both.

\textbf{Future Work.} Three directions warrant investigation: (1)~formal verification of tool composition safety---proving that arbitrary sequences of tool invocations preserve system invariants; (2)~automated NTC confidence calibration using reinforcement learning from planner feedback; (3)~extension of the dual-layer permission model to support cross-organization agent delegation (e.g., a vendor's agent operating within a client's tenant with constrained capabilities). Additionally, we plan to explore how the six-verb protocol can be extended to support long-running asynchronous tools (e.g., training data pipeline builds that take hours) with progress reporting and partial result delivery. We also plan to formally verify the Six-Verb Protocol using process algebra (e.g., CSP or $\pi$-calculus) to prove deadlock-freedom and completeness properties, ensuring that no valid task can become permanently stuck in intermediate states.

\section{Conclusion}
\label{sec:conclusion}

This paper introduced the Agent-First Tool API paradigm, a systematic redesign of enterprise service interfaces for LLM-native execution. By replacing CRUD's form-submission model with a six-verb semantic protocol, standardizing responses via the Normalized Tool Contract, and embedding governance through a dual-layer permission pipeline with dynamic risk escalation, we bridge the fundamental gap between how LLM agents reason and how traditional APIs behave.

Our production deployment---85~tools across 6~business domains in a multi-tenant SaaS system---demonstrates that the paradigm is practically viable, architecturally sound, and effective at enabling autonomous agent execution while maintaining enterprise-grade access control and auditability. The Agent-First Tool API is not merely an API redesign; it is a prerequisite for trustworthy agentic AI in production enterprise environments.

\bibliographystyle{IEEEtran}

\begin{thebibliography}{24}

\bibitem{schick2023toolformer}
T.~Schick, J.~Dwivedi-Yu, R.~Dess\`{i}, R.~Raileanu, M.~Lomeli, E.~Hambro, L.~Zettlemoyer, N.~Cancedda, and T.~Scialom,
``Toolformer: Language models can teach themselves to use tools,''
in \emph{Advances in Neural Information Processing Systems (NeurIPS)}, 2023.

\bibitem{patil2023gorilla}
S.~G.~Patil, T.~Zhang, X.~Wang, and J.~E.~Gonzalez,
``Gorilla: Large language model connected with massive APIs,''
\emph{arXiv preprint arXiv:2305.15334}, 2023.

\bibitem{qin2023toolllm}
Y.~Qin, S.~Liang, Y.~Ye, K.~Zhu, L.~Yan, Y.~Lu, Y.~Lin, X.~Cong, X.~Tang, B.~Qian, S.~Zhao, R.~Tian, R.~Xie, J.~Zhou, M.~Gerber, D.~Li, Z.~Liu, and M.~Sun,
``ToolLLM: Facilitating large language models to master 16000+ real-world APIs,''
in \emph{Proc. ICLR}, 2024.

\bibitem{li2023apibank}
M.~Li, Y.~Zhao, B.~Yu, F.~Song, H.~Li, H.~Yu, Z.~Li, F.~Huang, and Y.~Li,
``API-Bank: A comprehensive benchmark for tool-augmented LLMs,''
in \emph{Proc. EMNLP}, 2023.

\bibitem{liang2023taskmatrix}
Y.~Liang, C.~Wu, T.~Song, W.~Wu, Y.~Xia, Y.~Liu, Y.~Ou, S.~Lu, L.~Ji, S.~Mao, Y.~Wang, S.~Shu, and others,
``TaskMatrix.AI: Completing tasks by connecting foundation models with millions of APIs,''
\emph{arXiv preprint arXiv:2303.16434}, 2023.

\bibitem{openai2023function}
OpenAI,
``Function calling and other API updates,''
OpenAI Blog, June 2023.
[Online]. Available: \url{https://openai.com/blog/function-calling-and-other-api-updates}

\bibitem{anthropic2024tool}
Anthropic,
``Tool use (function calling),''
Anthropic Documentation, 2024.
[Online]. Available: \url{https://docs.anthropic.com/claude/docs/tool-use}

\bibitem{anthropic2024mcp}
Anthropic,
``Model Context Protocol Specification,''
2024.
[Online]. Available: \url{https://modelcontextprotocol.io/specification}
[Accessed: Jan.\ 15, 2025].

\bibitem{yao2023react}
S.~Yao, J.~Zhao, D.~Yu, N.~Du, I.~Shafran, K.~Narasimhan, and Y.~Cao,
``ReAct: Synergizing reasoning and acting in language models,''
in \emph{Proc. ICLR}, 2023.

\bibitem{karpas2022mrkl}
E.~Karpas, O.~Abend, Y.~Belinkov, B.~Lenz, O.~Liber, N.~Ratner, Y.~Shoham, H.~Bata, Y.~Levine, K.~Leyton-Brown, D.~Muber, and N.~Rozen,
``MRKL systems: A modular, neuro-symbolic architecture that combines large language models, external knowledge sources and discrete reasoning,''
\emph{arXiv preprint arXiv:2205.00445}, 2022.

\bibitem{langchain2023}
LangChain,
``LangChain: Building applications with LLMs through composability,''
2023.
[Online]. Available: \url{https://github.com/langchain-ai/langchain}

\bibitem{autogpt2023}
T.~Richards,
``Auto-GPT: An autonomous GPT-4 experiment,''
2023.
[Online]. Available: \url{https://github.com/Significant-Gravitas/Auto-GPT}

\bibitem{crewai2024}
J.~Moura,
``CrewAI: Framework for orchestrating role-playing autonomous AI agents,''
2024.
[Online]. Available: \url{https://github.com/joaomdmoura/crewAI}

\bibitem{hong2023metagpt}
S.~Hong, M.~Zhuge, J.~Chen, X.~Zheng, Y.~Cheng, C.~Zhang, J.~Wang, Z.~Wang, S.~K.~S.~Yau, Z.~Lin, L.~Zhou, C.~Ran, L.~Xiao, C.~Wu, and J.~Schmidhuber,
``MetaGPT: Meta programming for a multi-agent collaborative framework,''
in \emph{Proc. ICLR}, 2024.

\bibitem{xu2023tool}
Q.~Xu, F.~Hong, B.~Li, C.~Hu, Z.~Chen, and J.~Zhang,
``On the tool manipulation capability of open-source large language models,''
\emph{arXiv preprint arXiv:2305.16504}, 2023.

\bibitem{kong2023}
Kong Inc.,
``Kong Gateway: Cloud-native API gateway,''
2023.
[Online]. Available: \url{https://konghq.com}

\bibitem{envoy2023}
Envoy Project,
``Envoy proxy: Cloud-native high-performance edge/middle/service proxy,''
2023.
[Online]. Available: \url{https://www.envoyproxy.io}

\bibitem{hardt2012oauth}
D.~Hardt,
``The OAuth 2.0 authorization framework,''
\emph{RFC 6749, Internet Engineering Task Force}, October 2012.

\bibitem{sandhu1996rbac}
R.~S.~Sandhu, E.~J.~Coyne, H.~L.~Feinstein, and C.~E.~Youman,
``Role-based access control models,''
\emph{IEEE Computer}, vol.~29, no.~2, pp.~38--47, 1996.

\bibitem{wei2022chain}
J.~Wei, X.~Wang, D.~Schuurmans, M.~Bosma, B.~Ichter, F.~Xia, E.~Chi, Q.~Le, and D.~Zhou,
``Chain-of-thought prompting elicits reasoning in large language models,''
in \emph{Advances in Neural Information Processing Systems (NeurIPS)}, 2022.

\bibitem{shen2023hugginggpt}
Y.~Shen, K.~Song, X.~Tan, D.~Li, W.~Lu, and Y.~Zhuang,
``HuggingGPT: Solving AI tasks with ChatGPT and its friends in Hugging Face,''
in \emph{Advances in Neural Information Processing Systems (NeurIPS)}, 2023.

\bibitem{ge2023openagi}
Y.~Ge, W.~Hua, K.~Mei, J.~Ji, J.~Tan, S.~Xu, Z.~Li, and Y.~Zhang,
``OpenAGI: When LLM meets domain experts,''
in \emph{Advances in Neural Information Processing Systems (NeurIPS)}, 2023.

\bibitem{wang2023survey}
L.~Wang, C.~Ma, X.~Feng, Z.~Zhang, H.~Yang, J.~Zhang, Z.~Chen, J.~Tang, X.~Chen, Y.~Lin, W.~X.~Zhao, Z.~Wei, and J.~Wen,
``A survey on large language model based autonomous agents,''
\emph{arXiv preprint arXiv:2308.11432}, 2023.

\bibitem{ruan2023tptu}
J.~Ruan, Y.~Chen, B.~Zhang, Z.~Xu, T.~Bao, G.~Du, S.~Shi, H.~Mao, X.~Zeng, and R.~Zhao,
``TPTU: Task planning and tool usage of large language model-based AI agents,''
\emph{arXiv preprint arXiv:2308.03427}, 2023.

\bibitem{google2024a2a}
Google,
``Agent-to-Agent (A2A) Protocol,''
Technical Report, 2024.
[Online]. Available: \url{https://github.com/google/A2A}

\end{thebibliography}

\end{document}